\documentclass{amsart}
\usepackage[super,biblabel]{cite}
\usepackage{floatrow}
\usepackage[numbers]{natbib}
\usepackage{amsmath,amssymb,amsfonts}
\usepackage{graphicx}
\usepackage[foot]{amsaddr}
\usepackage{pslatex}
\usepackage{float} 

\begin{document}

\title[]{Generating Oscillation Activity with Echo State Network to Mimic the Behavior of \\ a Simple Central Pattern Generator}

\author{Tham Yik Foong$^{1}$}
\author{Danilo Vasconcellos Vargas$^{1,2}$}

\address{$^{1}$Kyushu University, Fukuoka, Japan}
\address{$^{2}$The University of Tokyo, Tokyo, Japan}

\maketitle

\begin{abstract}
This paper presents a method for reproducing a simple central pattern generator (CPG) using a modified Echo State Network (ESN). Conventionally, the dynamical reservoir needs to be damped to stabilize and preserve memory. However, we find that a reservoir that develops oscillatory activity without any external excitation can mimic the behaviour of a simple CPG in biological systems. We define the specific neuron ensemble required for generating oscillations in the reservoir and demonstrate how adjustments to the leaking rate, spectral radius, topology, and population size can increase the probability of reproducing these oscillations. The results of the experiments, conducted on the time series simulation tasks, demonstrate that the ESN is able to generate the desired waveform without any input. This approach offers a promising solution for the development of bio-inspired controllers for robotic systems.

\end{abstract}

\section{Introduction}


Central pattern generators (CPGs) are neural circuits (e.g., Stomatogastric ganglion circuit and crustacean cardiac ganglion) found in many animals that are responsible for generating rhythmic motor patterns, such as those used in walking, swimming, and breathing \cite{bib1, bib2, bib3, bib4}.
Most interestingly, such motor patterns can be achieved without a driving input.
Since its discovery, much research has focused on attempting to reproduce CPGs in order to understand how they work and potentially apply this knowledge to the development of, for instance, prosthetic limbs, robotic systems, and more \cite{bib3, bib5, bib6, bib8, bib9}. 
The researchers used a combination of artificial neural networks and mathematical modeling techniques to achieve this goal.


However, previous attempts to develop a computational model of CGPs have not been successful in creating a model that operates without any external input.
Mathematical models of CPGs have typically relied on predefined mathematical functions, such as the half-center model and Hopf oscillator, to act as pacemakers or half-center oscillators \cite{bib8, bib10, bib11, bib12, bib17}. 
Methods that used artificial neural networks to generate oscillation often involve networks with predefined topology, input, or external feedback to calibrate the output or generate oscillation \cite{bib9, bib13, bib14, bib15, bib16, bib27}, and in some cases, the model must be optimized before it can generate oscillation \cite{bib13, bib21}. 
In this study, we show how CPG can be reproduced with modified Echo State Networks (ESNs) to address these limitations.
The use of ESNs in this study is significant because they are known for their simplicity of implementation and their ability to model complex dynamics, making them well-suited for reproducing CPGs behaviour. 
Additionally, ESNs have the advantage of being able to work in an online mode, which means that the network can continue to learn and adapt as it receives new inputs.


That being said, this research aims to develop a simplified computational model of a CPG that can generate rhythmic patterns without any external input or feedback. 
The model, called Self-Oscillatory Echo State Network (SO-ESN), generates oscillations to reproduce desired waveforms.
The reproduced waveforms are identical to reproducing CPG motor patterns, which can be applied to the actuators of a robotic system to create rhythmic movement.
The novelty of our model lies in (1) its connectivity, which can generate spontaneous, diverse, and self-sustaining oscillations without any external input or feedback, (2) the model is simple and fast, and (3) the oscillation patterns are not predefined.


\section{Background}


\subsection{Echo State Network}

An ESN is a type of recurrent neural network (RNN) that is easy to train and excels in modeling dynamic systems \cite{bib19, bib20}. The network architecture consists of a `reservoir' where the connectivity and weights of reservoir units are randomly initialized, recurrent, and fixed throughout training. 
A readout layer is trained using the nonlinear response from the reservoir to predict a set of outputs. 
So far, ESNs have been successfully applied to a wide range of tasks, such as time series prediction, system identification, and natural language processing. 

Conventionally, the dynamical reservoir needs to be damped to preserve the memory of past inputs \cite{bib19}. 
In this paper, however, we discover that a reservoir can develop oscillatory activity without any external excitation, which mimics the basic characteristic of a CPG.

In this section, we briefly revisit the workflow of ESNs. 
In a classical ESN with $N$ neurons in the reservoir, the update is typically defined by a set of linear equations of the form:

\begin{equation}
x_{t+1} = (1-a) x_t + a \times f(W_{in}u_t + W  x_t)
\label{eq1}
\end{equation}

where $x_t$ is the state of the reservoir at time $t$, $u_t$ is the $K$ dimensional input, $W_{in}$ is the $N \times K$ weight matrix connecting the input to the reservoir, $W$ is the $N \times N$ weight matrix connecting the reservoir neurons to each other, $a$ is the leaking rate, and $f$ is a non-linear activation function. The future reservoir state $x_{t+1}$ is a function of the previous state $x_t$ and the input $u_t$ and is updated every time an input is presented. 

The output $y_t$ is generated by a linear combination of the current state of the reservoir $x_t$ and a weight matrix $W_{out}$. The equation for generating the output is typically defined as:

\begin{equation}
y_t = g(W_{out} x_t)
\label{eq2}
\end{equation}

where $y_t$ is the $L$ dimensional output at time $t$, $W_{out}$ is the $N \times L$ weight matrix connecting the reservoir to the output layer, and $g$ is an output activation function ($g$ can be omitted depending on the tasks). The only part of the network that is trained during the learning process is the weight matrix $W_{out}$, whereas the reservoir weights $W$ are fixed after initialization. 

\section{Self-Oscillatory Echo State Network (SO-ESN)}

Here, we introduce a SO-ESN consisting of a reservoir that generates oscillations without any feedback or input.
The weight and the initial state of the reservoir units are initialized arbitrarily from a uniform distribution [$-0.5, 0.5$].
Note that, the initial state of reservoir units should be non-zero to `kick start' the oscillation. 
The success rate for generating oscillations is not affected by the scaling of the initial state. 
However, a small initial state may delay the time it takes to reach the oscillation phase.
A readout layer is connected to the reservoir and is fine-tuned to produce the desired waves.

\begin{figure}[H]
\centerline{\includegraphics[width=0.6\linewidth]{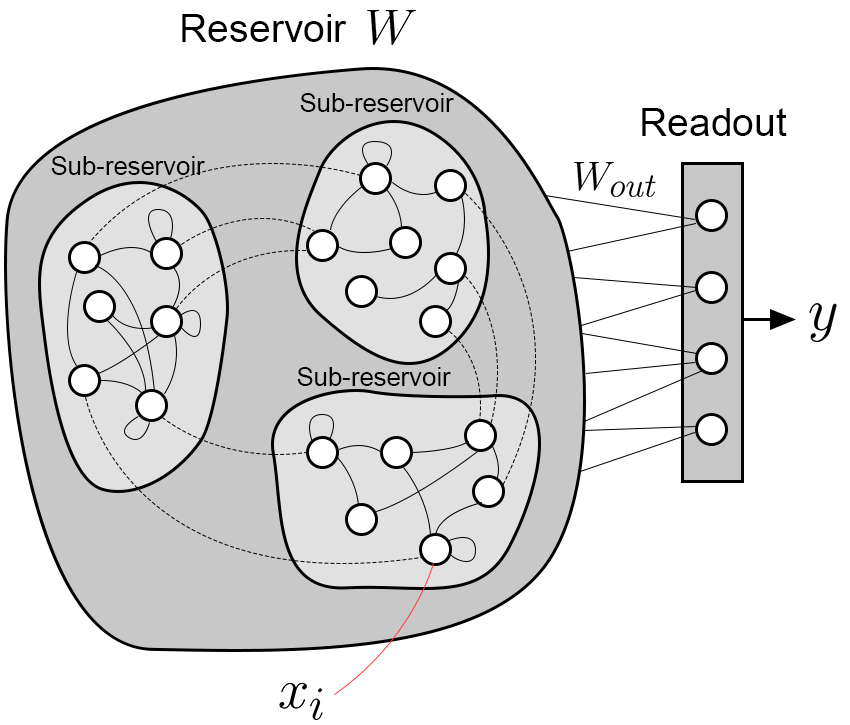}}
\caption{Model architecture of SO-ESN. $W$ and $W_{out}$ denote the connectivity of the reservoir and the reservoir to the readout layer, respectively, $x$ is the state of the reservoir units, and $y$ is the output.}
\label{fig1}
\end{figure}

Since our model architecture (Fig.~\ref{fig1}) does not contain an input layer or feedback weights, the form of the reservoir unit state update equation is modified:

\begin{equation}
x_{t+1} = (1-a) x_t + a \times f(W x_t)
\label{eq3}
\end{equation}

\noindent
Where $f$ is a $tanh$ function. 
Essentially, the current state of reservoir units is updated via the dot product of the weight matrix and the previous state, with respect to the leaking rate.
The leaking rate controls the extent to which the state of the reservoir neurons decays over time.
In other words, the leaking rate affects the memory capacity of the ESN.
A high leaking rate will make the network forget previous information quickly, while a low leaking rate will make the network retain previous information for a long time. 
Compared to the classical ESN, the leaking rate plays a different role in SO-ESN.
Here, it controls the frequency of the oscillation produced.

The output of SO-ESN is computed by Eq.~\ref{eq2}, but with the activation function $g$ omitted.
Ridge regression is used to train the readout layer $W_{out}$:

\begin{equation}
W_{out} = (X^\top X + \lambda I)^{-1}X^\top Y
\label{eq4}
\end{equation}

\noindent
where $X$ is the history of the reservoir units state $\left \{ x_0, x_1, \cdots ,x_{\tau} \right \} \in X$, $Y$ is the ground truth, and $\lambda = 1e-8$ is the ridge parameter. Note that Eq.~\ref{eq4} is not required to generate self-oscillation of the reservoir units.

Updating without any external input creates a scenario where the occurrence of oscillation highly relies on the connectivity, leaking rate, and spectral radius (the maximum absolute eigenvalue of the reservoir weight matrix) of the reservoir units.
These components are explored extensively in this paper.

\section{Results}

\subsection{Self-Oscillatory Reservoir can Generate Oscillations}


\begin{figure*}[htbp]
\centerline{\includegraphics[width=1\linewidth]{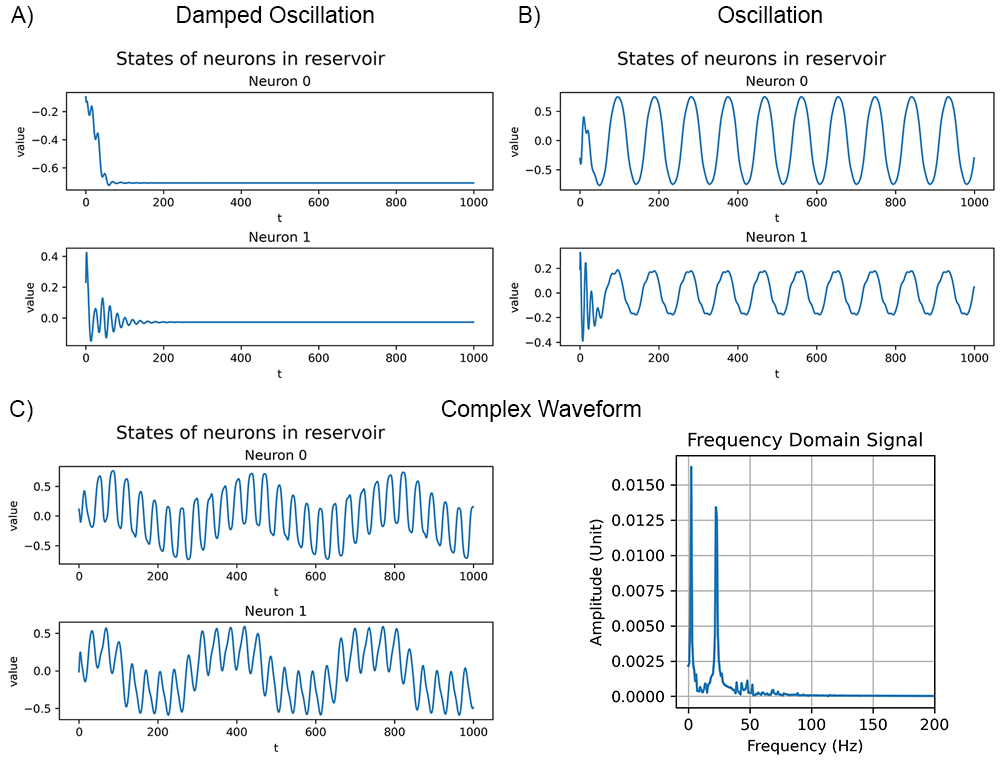}}
\caption{Examples of oscillations produced by the internal neurons of the self-oscillatory reservoirs. (A) Damped oscillation. (B) Self-sustaining oscillation generated from a self-oscillatory reservoir. After a few time steps, the initial conditions of the neurons are eliminated, and the neurons reach a steady oscillation phase. (C) Complex waveforms generated from a self-oscillatory reservoir. The analysis of complex waveforms using the fast Fourier transform reveals that they consist of multiple frequency components.
}
\label{fig2}
\end{figure*}

To understand the dynamics of our network, we randomly initialized several reservoirs and observed the reservoir units' states.
The reservoirs are initialized with a population of $N=100$, leaking rate $a = 0.5$, and spectral radius $\rho = 1.25$.
In Fig.~\ref{fig2}, we plot the dynamics of some reservoirs by running the networks for 1000 time steps and recording the state of the reservoir units.
The results illustrated that some reservoirs generate damped oscillations, where oscillations fade away and the state of the reservoir units converges to a constant value.
On the other hand, some reservoirs generate stable and self-sustaining oscillation.
We observed that the reservoir unit's state varies in amplitude but shares the same frequency. 
Such phase locking of frequency is induced by the strong coupling of reservoir units.
In this paper, we refer to the reservoir that oscillates as \textit{self-oscillatory reservoir}.


The self-oscillatory reservoir can generate complex waveforms, which are created spontaneously from the combination of multiple sub-sets of the coupled reservoir units.
These waveforms are important as they carry more information than real-valued waveforms, facilitating the reproduction of more complex behaviour.
Otherwise, we can obtain a complex waveform using the linear combination of the readout layer.

Interestingly, the self-oscillatory reservoir remains to obey the echo state property, where the initial state of the neurons is washed away after several time steps and enters a stable oscillation phase. 

\begin{figure*}[htbp]
\centerline{\includegraphics[width=1\linewidth]{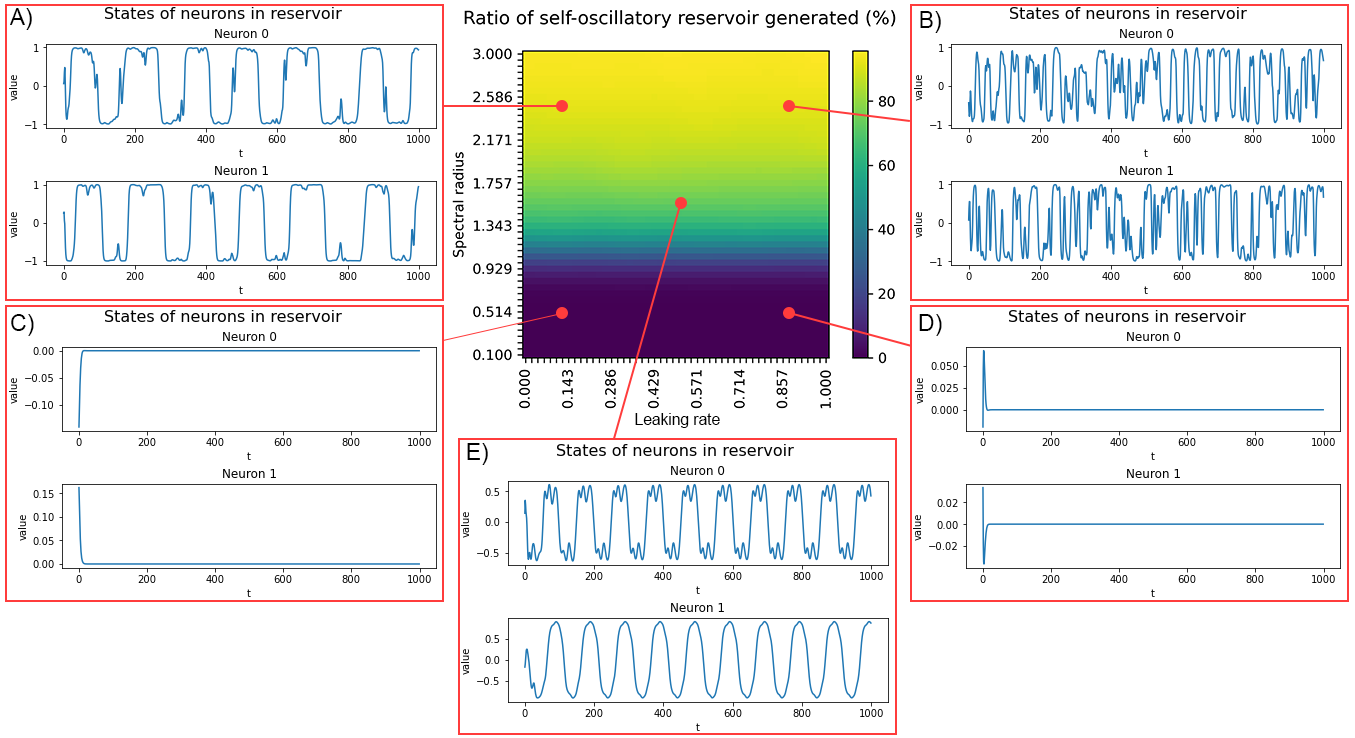}}
\caption{Ratio (\%) of self-oscillatory reservoir generated by tuning leaking rate and spectral radius. 
The heatmap indicates the ratio of self-oscillatory reservoirs ($N=100$) generated when adjusting the leaking rate and spectral radius.
(C, D) When the spectral radius is close to or less than one ($\rho<1$), the reservoir becomes less likely to self-oscillate.
(A, B) Besides, a larger spectral radius also creates a more chaotic oscillation than a smaller one (compare A, B with E).
On the other hand, the leaking rate does not affect the probability of generating a self-oscillatory reservoir;
However, it can adjust the oscillation frequency.
}
\label{fig3}
\end{figure*}

\subsection{Leaking Rate and Spectral Radius}

The findings in the previous subsection raised the questions of (1) what criteria are required for the reservoir to oscillate, (2) how to increase the probability of initializing a self-oscillatory reservoir. and (3) the effect of leaking rate and spectral radius on the produced oscillation.
To answer these questions, we performed the following experiment.
We perform parameter sweeping of the leaking rate and spectral radius with the range of $0 \leq a \leq 1$ and $0.1 \leq \rho \leq 3$ respectively.
With each set of parameters, we perform Monte Carlo sampling by generating 1000 reservoirs and determining the ratio of self-oscillatory reservoirs successfully generated.
We use the periodogram method to identify if the reservoir is self-oscillating by analyzing the signal produced in the last 100 timesteps of the reservoir units. 
This method checks if the spectral density of the signal has non-zero values at multiple frequency components, which indicates that the signal is likely oscillating.

The heatmap in Fig.~\ref{fig3} displays the relationship between the leaking rate and spectral radius of the self-oscillatory reservoir. 
The results demonstrated that when the spectral radius is close to or smaller than 1, the reservoir is less likely or unable to self-oscillate.
Additionally, a larger spectral radius tends to produce more chaotic oscillations than a smaller one. 
On the other hand, the leaking rate does not significantly impact the likelihood of generating a self-oscillatory reservoir, but it can be used to adjust the frequency of the oscillation.
Slow oscillation is more likely to be achieved when the leaking rate is small; 
Otherwise, a self-oscillatory reservoir tends to generate fast oscillation when the leaking rate is large.

\subsection{Neuron Ensemble Responsible for Oscillation}

\begin{figure}[htbp]
\centerline{\includegraphics[width=1\linewidth]{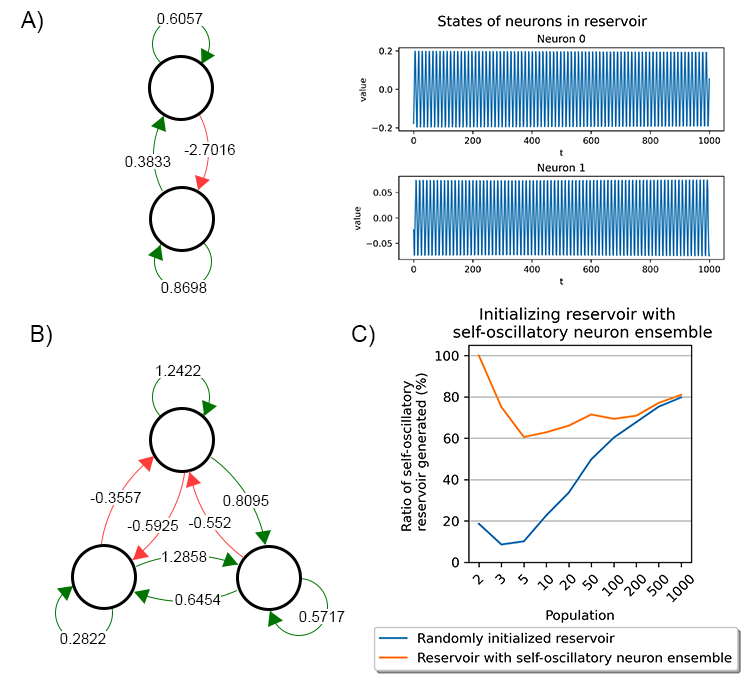}}
\caption{Examples of self-oscillatory neuron ensembles. Edges in green indicate excitatory synapses, and edges in red indicate inhibitory synapses. (A) Two reservoir units form the simplest form of a self-oscillatory neuron ensemble. The plot on the right shows the state of the two neurons' progress over time. 
(B) An instance of a self-oscillatory neuron ensemble composed of three reservoir units.
(C) The ratio of self-oscillatory reservoirs successfully generated with and without predefined self-oscillatory neuron ensembles from (A). 
}
\label{fig4}
\end{figure}

\begin{figure*}[htbp]
\centerline{\includegraphics[width=1\linewidth]{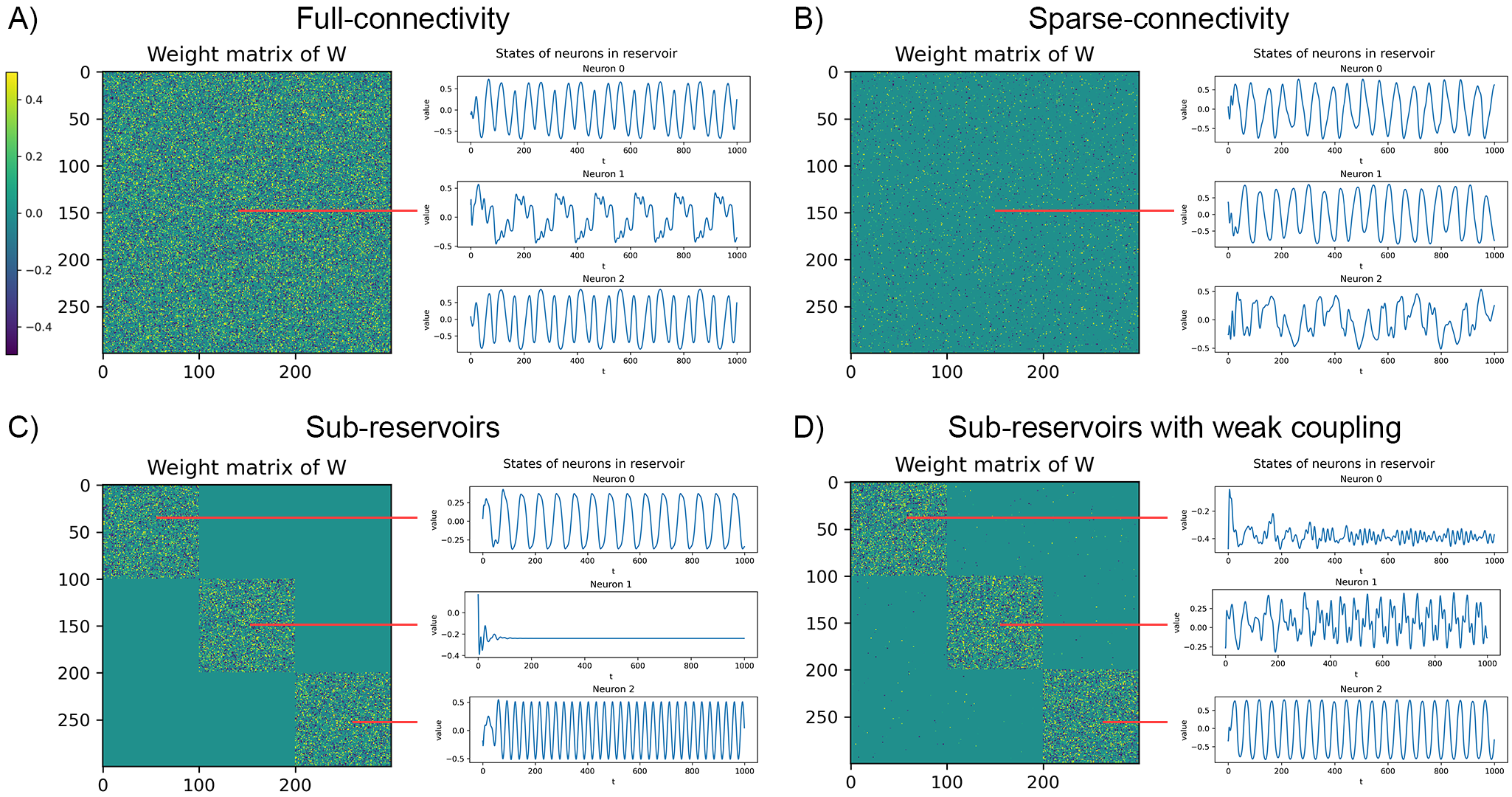}}
\caption{Weight matrix of the reservoirs with different topologies and the oscillation they generated. (A) A fully connected reservoir is only capable of producing oscillations at a single frequency. (B) A sparse connectivity reservoir generates oscillations at the same frequency. Although sparse connectivity does not directly decrease the likelihood of self-oscillatory reservoirs, the probability of generating low-frequency oscillations is limited by the population of reservoir units. (C) A reservoir composed of sub-reservoirs creates oscillations with various frequencies. Some sub-reservoirs may generate damped oscillation, which becomes a zero-variance feature for the readout layer. (D) Sub-reservoirs with weak coupling. Creating weak connections between sub-reservoirs increases the likelihood of the whole reservoir generating oscillations with various frequencies.}
\label{fig5}
\end{figure*}


In this section, we identify a few neuron ensembles responsible for generating oscillation within the reservoir, referred to as \textit{self-oscillatory neuron ensembles}.
Starting from the simplest form of a neuron ensemble, we showed (Fig.~\ref{fig4}A) that two neurons can oscillate via a reciprocal and recurrent connection. 
A neuron ensemble with such a configuration typically consists of 3 excitatory synapses and 1 inhibitory synapse.
In detail, neuron A `flips' the state value of connected neuron B, and then neuron B reciprocates by `flipping' the state value of neuron A.
This resembles the reciprocal inhibition often found in a biological CPG \cite{bib28}.

As the neuron ensemble population increases, the motif classes that can be formed also increase.
Thus, the degree of freedom in forming different types of signals also increases.
Based on empirical observation, neuron ensembles with larger populations can form more complex oscillations with more bands of frequency; 
But neuron ensembles with a smaller population only generate high-frequency oscillations.
The self-oscillatory neuron ensembles acting as a subset of the reservoir can couple with other non-oscillatory neuron ensembles and cause them to oscillate.
Krauss, Zankl, Schilling, Schulze, and Metzner conducted a review of analyses on network motifs, specifically focusing on a three-neuron structure identical to the one shown in Fig.~\ref{fig4}B \cite{bib29}.

In another experiment, we record the ratio of self-oscillatory reservoirs successfully generated with respect to the population (see Fig.~\ref{fig4}C).
Overall, the larger the population, the greater the chances of producing a self-oscillatory reservoir because there is a higher probability of having self-oscillatory neuron ensembles.
Further, we perform the same experiment but replace the first two neurons of the randomly initialized reservoir with the self-oscillatory neuron ensembles from Fig.~\ref{fig4}A.
We find that the probability of successfully generating a self-oscillatory reservoir becomes significantly higher.
Hence, this indicates that the main component needed to generate a self-oscillatory reservoir is to contain a self-oscillatory neuron ensemble.
Besides, the ratio grows slower as it approaches a higher population, converging with the result shown in the previous experiment.
This is due to the effect of the injected neuron ensemble becoming less significant.
Moreover, a small number of neurons added to a predefined self-oscillatory neuron ensemble can decrease the ratio of the self-oscillatory reservoirs generated, as a result of these additional neurons counteracting the neuron ensemble.
These analyses justified the conclusion that, when viewed from a network topological perspective, the interaction between reservoir units becomes an important building block for producing network-based rhythmicity.

\subsection{Weakly Coupled Sub-reservoirs Create Oscillation with Large Variation}


Up to this point, studies in the previous section only showed fully connected reservoirs, which are only able to generate phase-coupled oscillations.
This created nearly duplicate features (highly similar oscillation patterns), which provided limited contributions to the training of the readout layer.
Thus, a topology producing diverse features in SO-ESN dynamics is necessary.
Here, we explored the topologies of (1) sparse connectivity, (2) sub-reservoirs, and (3) sub-reservoirs with weak coupling and their impact on SO-ESN dynamics (Fig.~\ref{fig5}). 

Similar to the reservoir with full connectivity, a reservoir with sparse connectivity generates oscillations with the same frequency.
Although sparse connectivity does not directly contribute to the decrease in the probability of generating a self-oscillatory reservoir as long as self-oscillatory neuron ensembles exist.
But it is worth noting that the probability of generating low-frequency oscillations is bounded by the population of reservoir units.
Therefore, to generate a low-frequency oscillation with sparse connectivity, the population of the reservoir needs to be increased.

To generate oscillations with different frequencies, we can divide the reservoir into several sub-reservoirs (see Fig.~\ref{fig5}C).
The connectivity of such a reservoir can be described in the following block matrix form:

\begin{equation}
W = \begin{bmatrix}
\omega_0  & 0 & \cdots & 0\\ 
0 &  \omega_1 & \cdots & 0\\ 
\vdots  & \vdots &  \ddots & \vdots\\ 
0 & 0 & \cdots & \omega_M
\end{bmatrix}
\label{eq5}
\end{equation}

where the sub-reservoir is denoted as $\omega$ and $M$ as the total number of sub-reservoirs. 
To simplify, we defined the size of all sub-reservoirs as being constant.
These sub-reservoirs can be assigned with different leaking rates to create a variance in frequencies.
Instead of treating the leaking rate $a$ as a scalar in Eq.~\ref{eq3}, it can be represented as a leaking rate matrix $a \in A$ where each element is a random variable sampled from a Gaussian distribution. 
The elements of the $A$ can be represented by $a_{i,j} = N(\mu, \sigma), $ for $ i = 0, 1, ..., N $ and $ j = 0, 1, ..., N$.
Hence, the state update equation becomes $x_{t+1} = (1-A) x_t + A \times f(W x_t)$.
This topology helps create diverse features for the readout layer by providing oscillations with different frequencies.
However, some sub-reservoirs can generate damped oscillation, which becomes a zero-variance feature (features that do not provide any useful information) for the readout layer.

\begin{figure*}[htbp]
\centering
\includegraphics[width=1\linewidth]{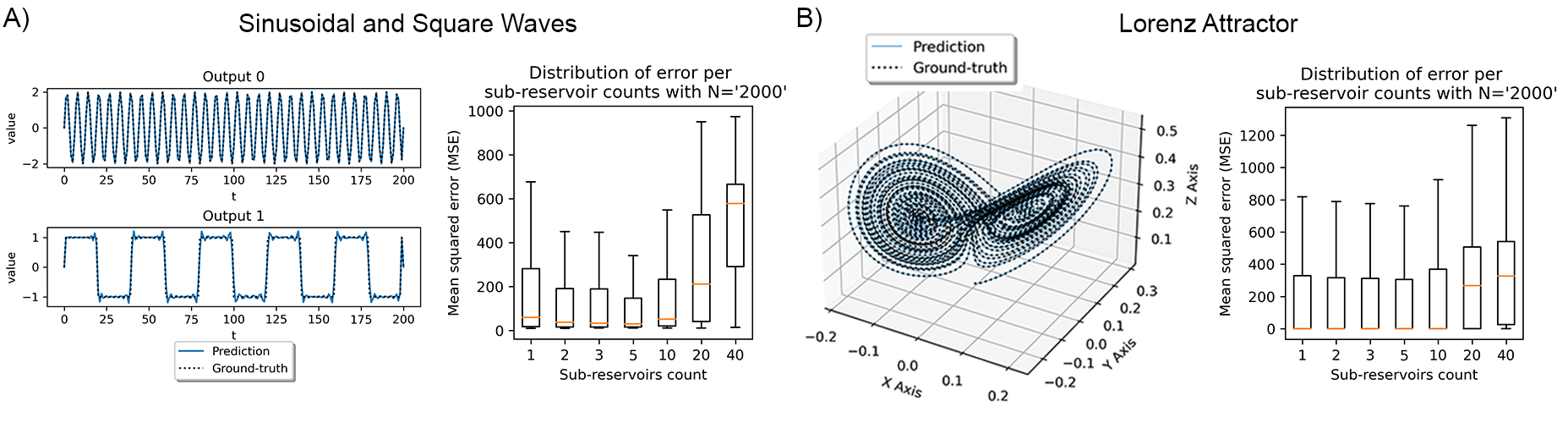}
\caption{SO-ESN reproduces the trajectory of complex dynamical systems. (A, B) SO-ESN can reproduce the trajectories of sinusoidal and square waves, and Lorenz chaotic time series with high precision, given oscillations with rich features produced by the self-oscillatory reservoir. The boxplot showed the optimal number of sub-reservoirs for a fixed population for both experiments.}
\label{fig6}
\end{figure*}

To make this better, sparse connections can be established between the sub-reservoirs to form coupled oscillators.
This approach allows sub-reservoirs that are capable of self-oscillation to transfer their oscillatory behaviour to sub-reservoirs that are not self-oscillatory.
This way, the entire reservoir dynamics can be decomposed into loosely coupled subsystems that generate oscillation with large variations among the sub-reservoirs.
The connectivity of such a topology can be described by:

\begin{equation}
W = \begin{bmatrix}
\omega_0  & \omega_{1,0} & \cdots & \omega_{M,0}\\ 
\omega_{0,1} &  \omega_1 & \cdots & \omega_{M,1}\\ 
\vdots  & \vdots &  \ddots & \vdots\\ 
\omega_{0,M} & \omega_{1,M} & \cdots & \omega_M
\end{bmatrix}
\label{eq6}
\end{equation}

As a result, the weakly coupled sub-reservoir's generated oscillations provide a diverse set of features that aid in training the readout layer.

\subsection{Reproducing Complex Dynamic Systems}


To demonstrate that our model can perform basic CPG behaviour without any external input, we conduct experiments reproducing the trajectory of complex dynamic systems using SO-ESN.
The goal of SO-ESN is to replicate the trajectories by tuning the readout layer with the actual trajectory, utilizing merely the state of reservoir units.

Firstly, we generate the ground truth sinusoidal and square waves, defined as:

\begin{equation}
\frac{dx}{dt} = 2(1 + cos(t)), y(t) = sgn(sin 10 \pi t)
\end{equation}

with $sgn$ be the sign function.
In the second experiment, we increase the complexity of the problem to reproduce the Lorenz attractor time series with SO-ESN:

\begin{equation} 
\frac{dx}{dt} = \sigma(y-x), \frac{dy}{dt} = x(\alpha-z) - y, \frac{dz}{dt} = xy - \beta z
\end{equation}

with $x(0)=0$, $y(0)=1$, and $z(0)=1.05$ as the initial state variables and $\sigma=10$, $\alpha=28$, and $\beta=2.667$ as the system parameters.

In both experiments, we set the population size of SO-ESN to $N=2000$ with the total time steps $\tau = 1000$ for the sinusoidal and square waves problem and $\tau = 5000$ for the Lorenz attractor time series problem.
The leaking rate matrix $A$ is drawn from a Gaussian distribution $N(0.6, 0.1)$ and we set the spectral radius to $\rho = 1.25$.
Different values of hyperparameters are tested heuristically, where the above setting provides an optimum result with an ideal computational time.
Both experiments are performed with 1000 trials, and a new reservoir is initiated for each new trial.


Fig~\ref{fig6} demonstrated that the SO-ESN effectively produced accurate reproduction of waveforms in both experiments. 
However, if the reservoir is not oscillating or has many nearly duplicate features, the linear combination of the reservoir units' states and output weights will not be able to reproduce the waveforms accurately. 
The boxplot in Fig~\ref{fig6} also illustrates that an optimal number of sub-reservoirs is necessary for a constant population, as too few sub-reservoirs will not have the ability to generate oscillations with spatially distinct features. 
Similarly, when the population is distributed among many sub-reservoirs, the sub-reservoirs will be too small to generate oscillations.

\section{Discussion}


In this study, we developed a model capable of generating spontaneous, self-sustaining rhythmic oscillations with diverse waveforms without external input or feedback. 
We discovered that the key to this model's ability to generate oscillations is the presence of self-oscillatory neuron ensembles.


Currently, although we found a way to increase the likelihood of a self-oscillatory reservoir through random initialization, a self-oscillatory reservoir cannot be guaranteed.
An important direction of future work should focus on finding a method that ensures a self-oscillating reservoir with the desired phase, frequency, and amplitude required. 
This could potentially be achieved through the use of evolutionary algorithms or other methods \cite{bib23, bib24} to modify the reservoir's weights and connections.

Another extended work is to further approximate the functionality of a biological CPG. 
This includes being able to control the gait transition or to activate, modify, and deactivate the CPG circuit through the use of neuromodulators \cite{bib2}.
In this context, the SO-ESN could function as a circuit that receives high-level commands to respond to environmental perturbations \cite{bib25, bib8}.
Such commands should modulate the characteristics of the generated oscillations that change the output adaptively.


To sum up, our model demonstrated the ability to reproduce motor patterns akin to simple CPG behaviours and its simplicity might offer a promising path to developing biologically inspired robotic controllers.

\section{Acknowledgments}
This work was supported by JST, ACT-I Grant Number JP-50243 and JSPS KAKENHI Grant Number JP20241216. T.Y.F. is supported by JST SPRING, Grant Number JPMJSP2136.

\typeout{}
\bibliographystyle{unsrt}
\bibliography{main}

\begin{thebibliography}{10}

\bibitem{bib1}
Eve Marder and Dirk Bucher.
\newblock Central pattern generators and the control of rhythmic movements.
\newblock {\em Current biology}, 11(23):R986--R996, 2001.

\bibitem{bib2}
Ronald~M Harris-Warrick.
\newblock Neuromodulation and flexibility in central pattern generator
  networks.
\newblock {\em Current opinion in neurobiology}, 21(5):685--692, 2011.

\bibitem{bib3}
Auke~Jan Ijspeert.
\newblock Central pattern generators for locomotion control in animals and
  robots: a review.
\newblock {\em Neural networks}, 21(4):642--653, 2008.

\bibitem{bib4}
David~W Sims, Nicolas~E Humphries, Nan Hu, Violeta Medan, and Jimena Berni.
\newblock Optimal searching behaviour generated intrinsically by the central
  pattern generator for locomotion.
\newblock {\em Elife}, 8:e50316, 2019.

\bibitem{bib5}
Auke~Jan Ijspeert and Alessandro Crespi.
\newblock Online trajectory generation in an amphibious snake robot using a
  lamprey-like central pattern generator model.
\newblock In {\em Proceedings 2007 IEEE International Conference on Robotics
  and Automation}, pages 262--268. IEEE, 2007.

\bibitem{bib6}
Rafael~R Torrealba, Jos{\'e} Cappelletto, Leonardo Ferm{\'\i}n, Gerardo
  Fern{\'a}ndez-L{\'o}pez, and Juan~C Grieco.
\newblock Cybernetic knee prosthesis: application of an adaptive central
  pattern generator.
\newblock {\em Kybernetes}, 41(1/2):192--205, 2012.

\bibitem{bib8}
Guillaume Sartoretti, Samuel Shaw, Katie Lam, Naixin Fan, Matthew Travers, and
  Howie Choset.
\newblock Central pattern generator with inertial feedback for stable
  locomotion and climbing in unstructured terrain.
\newblock In {\em 2018 IEEE international conference on robotics and automation
  (ICRA)}, pages 5769--5775. IEEE, 2018.

\bibitem{bib9}
Pablo Lopez-Osorio, Alberto Patino-Saucedo, Juan~P Dominguez-Morales, Horacio
  Rostro-Gonzalez, and Fernando Perez-Pe{\~n}a.
\newblock Neuromorphic adaptive spiking cpg towards bio-inspired locomotion.
\newblock {\em Neurocomputing}, 502:57--70, 2022.

\bibitem{bib10}
Alessandro Crespi and Auke~Jan Ijspeert.
\newblock Amphibot ii: An amphibious snake robot that crawls and swims using a
  central pattern generator.
\newblock In {\em Proceedings of the 9th international conference on climbing
  and walking robots (CLAWAR 2006)}, number CONF, pages 19--27, 2006.

\bibitem{bib11}
Dai Owaki and Akio Ishiguro.
\newblock A quadruped robot exhibiting spontaneous gait transitions from
  walking to trotting to galloping.
\newblock {\em Scientific reports}, 7(1):1--10, 2017.

\bibitem{bib12}
Haitao Yu, Haibo Gao, and Zongquan Deng.
\newblock Enhancing adaptability with local reactive behaviors for hexapod
  walking robot via sensory feedback integrated central pattern generator.
\newblock {\em Robotics and Autonomous Systems}, 124:103401, 2020.

\bibitem{bib17}
Shinya Aoi, Yoshimasa Egi, Ryuichi Sugimoto, Tsuyoshi Yamashita, Soichiro
  Fujiki, and Kazuo Tsuchiya.
\newblock Functional roles of phase resetting in the gait transition of a biped
  robot from quadrupedal to bipedal locomotion.
\newblock {\em IEEE Transactions on Robotics}, 28(6):1244--1259, 2012.

\bibitem{bib13}
Francis Wyffels and Benjamin Schrauwen.
\newblock Design of a central pattern generator using reservoir computing for
  learning human motion.
\newblock In {\em 2009 Advanced Technologies for Enhanced Quality of Life},
  pages 118--122. IEEE, 2009.

\bibitem{bib14}
Beck Strohmer, Poramate Manoonpong, and Leon~Bonde Larsen.
\newblock Flexible spiking cpgs for online manipulation during hexapod walking.
\newblock {\em Frontiers in neurorobotics}, 14:41, 2020.

\bibitem{bib15}
Horacio Rostro-Gonzalez, Pedro~Alberto Cerna-Garcia, Gerardo Trejo-Caballero,
  Carlos~H Garcia-Capulin, Mario~Alberto Ibarra-Manzano, Juan~Gabriel
  Avina-Cervantes, and C{\'e}sar Torres-Huitzil.
\newblock A cpg system based on spiking neurons for hexapod robot locomotion.
\newblock {\em Neurocomputing}, 170:47--54, 2015.

\bibitem{bib16}
Alexander Vandesompele, Gabriel Urbain, Joni Dambre, et~al.
\newblock Populations of spiking neurons for reservoir computing: Closed loop
  control of a compliant quadruped.
\newblock {\em Cognitive Systems Research}, 58:317--323, 2019.

\bibitem{bib27}
Jiwen Li.
\newblock Minimal energy control of an esn pattern generator.
\newblock 2011.

\bibitem{bib21}
T~Konstantin Rusch and Siddhartha Mishra.
\newblock Coupled oscillatory recurrent neural network (cornn): An accurate and
  (gradient) stable architecture for learning long time dependencies.
\newblock {\em arXiv preprint arXiv:2010.00951}, 2020.

\bibitem{bib19}
Herbert Jaeger.
\newblock Tutorial on training recurrent neural networks, covering bppt, rtrl,
  ekf and the" echo state network" approach.
\newblock 2002.

\bibitem{bib20}
Herbert Jaeger.
\newblock The “echo state” approach to analysing and training recurrent
  neural networks-with an erratum note.
\newblock {\em Bonn, Germany: German National Research Center for Information
  Technology GMD Technical Report}, 148(34):13, 2001.

\bibitem{bib28}
Dirk Bucher, Gal Haspel, Jorge Golowasch, and Farzan Nadim.
\newblock Central pattern generators.
\newblock {\em eLS}, pages 1--12.

\bibitem{bib29}
Patrick Krauss, Alexandra Zankl, Achim Schilling, Holger Schulze, and Claus
  Metzner.
\newblock Analysis of structure and dynamics in three-neuron motifs.
\newblock {\em Frontiers in Computational Neuroscience}, 13:5, 2019.

\bibitem{bib23}
Chang Liu, Jia-Qi Dong, Qing-Jian Chen, Zi-Gang Huang, Liang Huang, Hai-Jun
  Zhou, and Ying-Cheng Lai.
\newblock Controlled generation of self-sustained oscillations in complex
  artificial neural networks.
\newblock {\em Chaos: An Interdisciplinary Journal of Nonlinear Science},
  31(11):113127, 2021.

\bibitem{bib24}
Zhi-Hui Zhan, Jian-Yu Li, and Jun Zhang.
\newblock Evolutionary deep learning: A survey.
\newblock {\em Neurocomputing}, 483:42--58, 2022.

\bibitem{bib25}
S{\'e}bastien Gay, Jos{\'e} Santos-Victor, and Auke Ijspeert.
\newblock Learning robot gait stability using neural networks as sensory
  feedback function for central pattern generators.
\newblock In {\em 2013 IEEE/RSJ international conference on intelligent robots
  and systems}, pages 194--201. Ieee, 2013.

\end{thebibliography}

\end{document}